# Context-Emotion Aware Therapeutic Dialogue Generation: A Multi-component Reinforcement Learning Approach to Large Language Models for Mental Health Support

Eric H.Q. Zhang and Julia Ive

University College London, London, United Kingdom
eric.zhang.24@ucl.ac.uk, j.ive@ucl.ac.uk

**Abstract.** Mental health disorders impose a substantial global socioeconomic burden, with COVID-19 exacerbating mental health service accessibility challenges and boosting demand for telehealth support. While large language models (LLMs) offer 24/7, non-judgmental interactions to address this gap, pretrained models lack contextual coherence and emotional alignment for appropriate therapeutic dialogue. Existing methods suffer from three critical methodological gaps: 1) Supervised Fine-Tuning (SFT) produces repetitive, context-insensitive outputs that fail to balance clinical accuracy with genuine empathy; 2) Reinforcement Learning (RL)-based therapeutic systems rely on generic reward functions (e.g., BLEU, ROUGE) that prioritise lexical similarity over clinical-specific emotional appropriateness and contextual relevance; 3) LLMs are resource-intensive and pose data privacy risks, making local deployment in clinical settings infeasible. To address these gaps, this study investigates the application of SFT and RL techniques to enhance GPT-2's capacity for therapeutic dialogue generation. The methodology restructured input formats to enable simultaneous processing of contextual information and emotional states alongside user input, employing a novel multi-component reward function that explicitly aligns model outputs with professional therapeutic logic (not just lexical overlap) and annotated emotions. Results demonstrated substantial improvements through RLs over baseline GPT-2 across multiple evaluation metrics: BLEU (0.0111), ROUGE-1 (0.1397), ROUGE-2 (0.0213), ROUGE-L (0.1317), and METEOR (0.0581). LLM evaluation confirmed high contextual relevance and professionalism, while RL achieved 99.34% emotion accuracy compared to 66.96% for baseline GPT-2. These findings demonstrate RL's effectiveness in developing therapeutic dialogue systems that can serve as valuable assistive tools for therapists, while maintaining essential human clinical oversight. Code & Appendix: https://github.com/ez-anthro-tech-design/RLforGPT2MentalHealth

## 1 Introduction

Mental illness imposes a significant socioeconomic burden, with one in five adults in the United States experiencing a mental health condition each year [1]. Patients are further deterred from seeking care due to societal stigma [2], attitudinal barriers [3], and high costs and lack of knowledge about accessing these services [4].

Direct patient-facing AI systems without clinical supervision face significant obstacles, as patients demonstrate a lack of trust and lower adherence towards autonomous non-human consultations [5]. Concerns exist regarding patient safety and accuracy when AI-assisted dialogue tools operate without clinical oversight [6], while limited laws governing ethical use, data privacy, and transparent communication in autonomous AI systems compound these issues [7]. Additionally, LLMs exhibit hallucination and may be pretrained on datasets of unknown validation status, creating liability uncertainties on AI-generated clinical guidance as standard-of-care evidence despite the absence of expert review protocols [8]. Given these constraints, LLMs may be most effectively deployed as supervised dialogue generation tools that support healthcare professionals across various therapeutic contexts – including telehealth and in-person sessions – by providing real-time conversational suggestions rather than replacing clinical judgment.

Beyond these deployment challenges, technical limitations in existing fine-tuning methods create critical methodological gaps for therapeutic dialogue generation:

1. SFT operates on a supervised learning paradigm, where pre-trained models learn from input-output pairs via cross-entropy loss minimisation, updating parameters to reduce prediction errors on labelled datasets [9]. However, Stiennon et al. [10] showed that SFT is limited in penalising equally critical errors and minor lexical variations, risking reinforcing biases in low-quality training data, and generating repetitive outputs. SFT struggles to align with human therapeutic preferences [11] [12] – patients prioritise genuine empathy over technically correct but emotionally shallow responses [13], yet SFT models lack mechanisms to integrate affective empathy with clinical logic.
2. RL offers a robust approach to align model outputs with human preferences through human preference collection, reward model learning, and policy optimisation [14]. RL learns via environmental interactions, using reward signals to refine policies through techniques like proximal policy optimisation (PPO) [18]. Existing RL-based therapeutic systems (e.g., [15] [16]) use reward functions measured by generic NLP metrics (BLEU, ROUGE-L) or surface-level sentiment scores. These metrics prioritise lexical overlap with reference texts over clinical relevance, resulting in responses that are grammatically consistent but fail to address patient-specific contexts or align with therapeutic emotional boundaries.

In this study, we present three key contributions that directly address the limitations of existing methods:

1. **A clinical-specific multi-component reward function**: Unlike generic reward functions (e.g., [16]) that prioritise lexical similarity, our reward function integrates five interdependent clinical dimensions-text quality (professionalism), emotional alignment (therapeutic boundary-aware), contextual relevance (patient-specific), empathetic content (genuine understanding), and sentiment appropriateness (supportive tone). This explicitly solves the problem of "reward hacking" (e.g., generic empathy/ repetitive phrases) and ensures responses meet clinical standards, not just NLP metrics.

2. **An RL framework for affective empathy enhancement**: Our framework builds on SFT to model context and emotion as a unified input (via structured tokens), rather than separate features. This addresses the context-emotion modeling deficit of prior work, enabling the model to generate emotionally appropriate responses that are tied to patient context. Restructuring GPT-2's structure allows for explicit generation of emotion tokens alongside text, providing therapists with transparent examples of ideal therapeutic communication (context + emotion) for supervised use and clear demonstrations of ideal therapeutic communication for use across supervised clinical settings.
3. **A lightweight, locally deployable implementation**: By leveraging GPT-2 (124M parameters), we avoid the resource and privacy barriers of large LLMs. GPT-2's open-source nature and low computational requirements enable easy deployment, solving the accessibility gap for clinics with limited resources.

Our results demonstrated substantial improvements through reinforcement learning over baseline GPT-2 across multiple evaluation metrics: BLEU improved by 428% (0.0021 to 0.0111), ROUGE-1 by 192% (0.0478 to 0.1397), ROUGE-2 by 233% (0.0064 to 0.0213), ROUGE-L by 223% (0.0408 to 0.1317). LLM-as-a-judge confirmed high contextual relevance and professionalism, whilst reinforcement learning achieved 99.34% emotion accuracy, representing a 48% improvement over baseline GPT-2's 66.96%.

The remainder of this study is structured as follows: Section 2 presents the related work; Section 3 introduces the dataset and research questions; Section 4 details the methodology, including the multi-component reward function and RL framework; Section 5 presents the experimental results; Section 6 discusses the findings and contextualises them with related work; and Section 7 concludes with key contributions and future directions. Finally, Section 8 details the ethics.

## 2 Related Work

### 2.1 AI-Assisted Therapeutic Dialogue Tools in Clinical Practice

The integration of AI-assisted dialogue tools in mental health services has emerged as a promising approach to enhance therapeutic communication while maintaining necessary clinical supervision. Telehealth services have demonstrated effectiveness as scalable and cost-effective alternatives to traditional face-to-face mental health services, helping to overcome barriers such as social stigma, limited accessibility, and high costs [17] [18]. Similarly, text-based therapeutic interventions have also shown promise, with email-based exercises incorporating positive psychology interventions demonstrating improvements in stress levels and psychological well-being [19], indicating that text-based digital therapeutic communication can effectively support psychological health.

AI-assisted dialogue generation tools, when integrated into supervised therapeutic settings, offer potential solutions to enhance communication quality while addressing these limitations. These tools can support therapists by providing contextually appropriate dialogue suggestions in real-time, whether during telehealth sessions or in-person consultations.

Studies have demonstrated the effectiveness of LLM-backed therapeutic support tools, which are valued for their non-judgmental nature and ability to provide structured conversational frameworks that enhance patient comfort [20]. When used under clinical supervision, these tools can offer consistent dialogue suggestions and emotional support frameworks during periods when therapists need alternative phrasing options or guidance on addressing sensitive topics [20]. The key distinction lies in positioning these tools as augmentative rather than replacement technologies, ensuring that clinical judgment and human empathy remain central to therapeutic practice while leveraging AI capabilities to enhance communication effectiveness.

### 2.2 Fine-Tuning LLMs for Mental Health Support

However, existing tools fail to address core clinical needs. Much of the literature exploring the use of pre-trained LLMs for mental health support (MHS) relies heavily on prompt engineering approaches. For example, Gabriel et al. [21] found that GPT-4 can generate responses with nearly 45% greater overall empathy than human peer-to-peer responses. This contrasts with human therapists, who balance empathy with neutrality to avoid overwhelming patients [22]

Beyond prompt engineering, supervised fine-tuning approaches have shown significant promise [23]. Jin et al. [24] demonstrated that fine-tuned models outperform pre-trained models on emotional support tasks, yet argued that fine-tuned models struggle with human-like emotional support and risk producing empty, repetitive outputs. These findings highlight the need for more sophisticated training approaches that go beyond traditional supervised learning.

RL has emerged as a promising complementary approach to address these limitations. Sharma et al. [15] applied RL to DialoGPT, a large dialogue generation model based on GPT-2, for empathic rewriting, aiming to transform low-empathy conversational posts to higher empathy responses. While Sharma et al.'s [15] work demonstrated the feasibility of RL in improving response generation, achieving a BLEU score of 0.1391, they focused on "empathic rewriting" (improving existing low-empathy posts) rather than generative dialogue. Their reward function prioritises paraphrasing quality, not clinical appropriateness – making it irrelevant for real-time therapeutic interactions. Similarly, Saha et al. [16] employed RL to fine-tune GPT-2 [25] for empathetic text generation using a combined reward function including BLEU, ROUGE-L, and sentiment scores. However, their focus on lexical and sentiment metrics means their model cannot distinguish between contextually relevant empathy. For example, a patient with PTSD experiencing flashbacks requires neutral, grounding responses [22], but existing RL-fine-tuned small LM models (e.g., [16]) would reward overly empathetic responses (e.g., "That must be terrifying – you're so brave") that may exacerbate hyperarousal. These studies primarily focused on general empathy metrics rather than clinically specific emotional appropriateness and contextual relevance, indicating a gap that our work aims to address.

# 3 Dataset

## 3.1 Training Dataset

**Table 1.** Dialogue characteristics of the MESC dataset

| | **Total Records** | **Unique Problem Types** | **Min. Dialogue Turns** | **Max. Dialogue Turns** | **Avg. Dialogue Turns** |
|---|---|---|---|---|---|
| Train | 815 | 15 | 8 | 99 | 28.4 |
| Val | 102 | 15 | 11 | 47 | 26.6 |
| Test | 102 | 15 | 9 | 57 | 28.6 |

Table 1 describes the structure of the MESC dataset [26], which was selected to address the context-emotion modeling deficit of prior work. Notably, this dataset captures multimodal cues – including spoken dialogue, facial expressions, and body language – alongside text, thereby addressing a critical limitation of earlier text-only emotional support datasets. The dataset is adapted from the TV series *In Treatment*, which features professional actors portraying therapists and patients in realistic therapy sessions. The performances were validated by mental health experts at the Shanghai Mental Health Centre to ensure alignment with genuine therapeutic interactions [26].

The dataset includes seven emotion categories (anger, sadness, depression, disgust, fear, joy, and neutral) with labels calibrated by GPT-3.5 and manually validated by emotional support researchers. Fig. 1 shows the emotion distribution, in which neutral emotions are dominant for therapists – a reflection of clinical best practices [22] that guided our reward function's emphasis on emotion alignment.

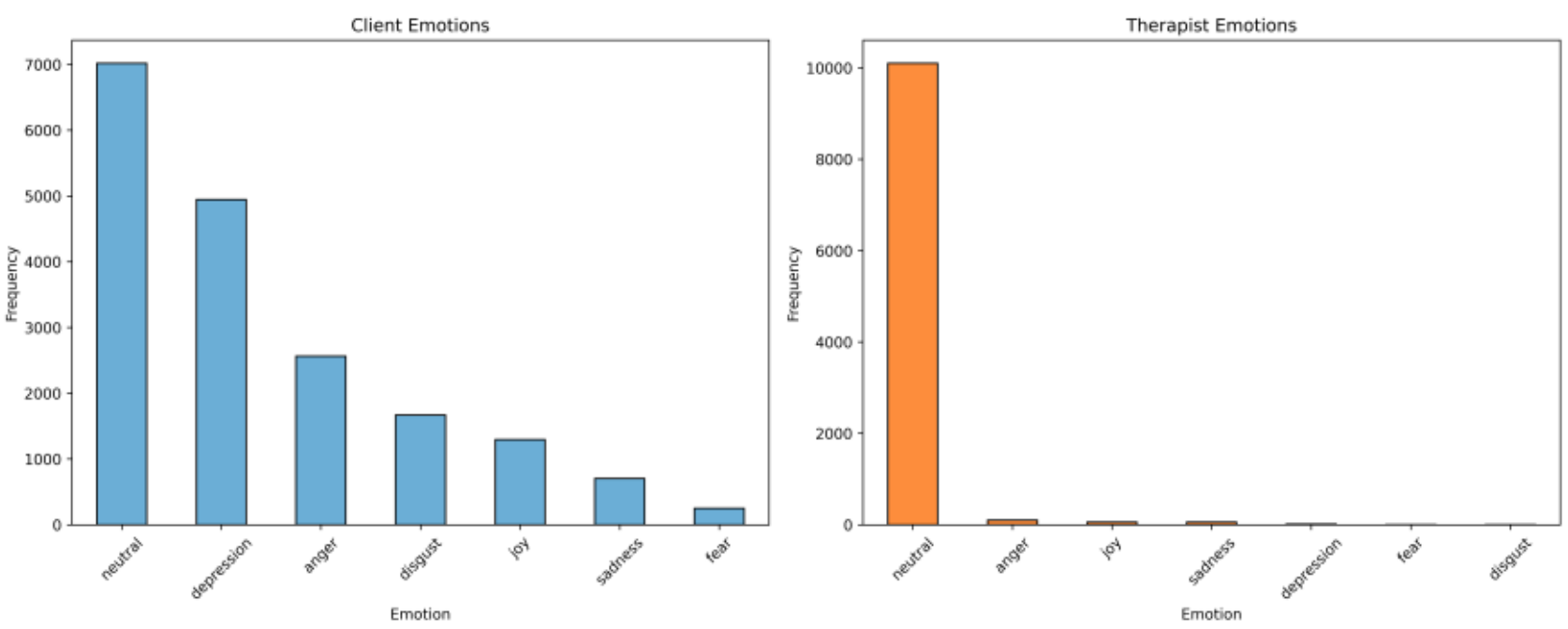

**Fig. 1.** Distribution of emotion categories by speaker (MESC dataset).

### 3.2 Evaluation Dataset

The ESConv dataset is a high-quality collection of emotional support dialogues between “help-seekers” (simulating patients) and “supporters” (simulating therapists), with rich metadata, including emotion types, problem categories, and support strategies [27]. It was chosen to test generalizability beyond MESC’s therapy-specific scenarios. As shown in Table 2, it contains 1300 annotated dialogues with problem categories and support strategies, providing a complementary context of “emotional support”. To align with our context-emotion framework, we used the roberta-base-go_emotions model to classify ESConv’s utterances into MESC’s 7 emotion categories. ESConv’s dialogues focus on “emotional support” (rather than MESC’s “therapeutic sessions”) - a related but distinct context. This ensures the model’s improvements are not limited to the training data’s specific scenario but apply to broader mental health support contexts.

**Table 2.** Dialogue characteristics of the ESConv dataset

| Metric | Value |
|---|---|
| Total conversations | 1300 |
| Minimum dialogue turns | 16 |
| Maximum dialogue turns | 120 |
| Mean dialogue turns | 29.51 |
| Median dialogue turns | 27.0 |
| Mean words per turn | 16.4 |

### 3.3 Data Preprocessing

During preprocessing, text extraction isolated authentic speech from metadata descriptions using regular expressions to remove patterns beginning with “The speaker” or “The emotion state”. Utterances were then concatenated before changing speakers.

**Sequence Filtering.** A 128-token threshold (see Fig. 2) was selected to preserve computational efficiency while retaining ~95% of data. This ensures that the model processes complete contextual-emotional units without excessive padding.

**Tokenizer Setup and Vocabulary Extension.** This study extended the original GPT-2 tokenizer (~50,257 tokens) with custom special tokens (<problem>, <user>, <user_emotion>, <therapist>, <therapist_emotion>) and seven emotion tokens ([“anger”, “sadness”, “depression”, “disgust”, “fear”, “joy”, “neutral”]). This structure (Table 3) enables delineation of dialogue components, allowing the model to distinguish between patient context, utterances, and emotional states. Emotion tokens are treated as atomic units to preserve semantic integrity – otherwise, subword tokenisation might fragment “sadness” into [“sad”, “ness”]. Crucially, it also allows explicit generation of emotion tokens.

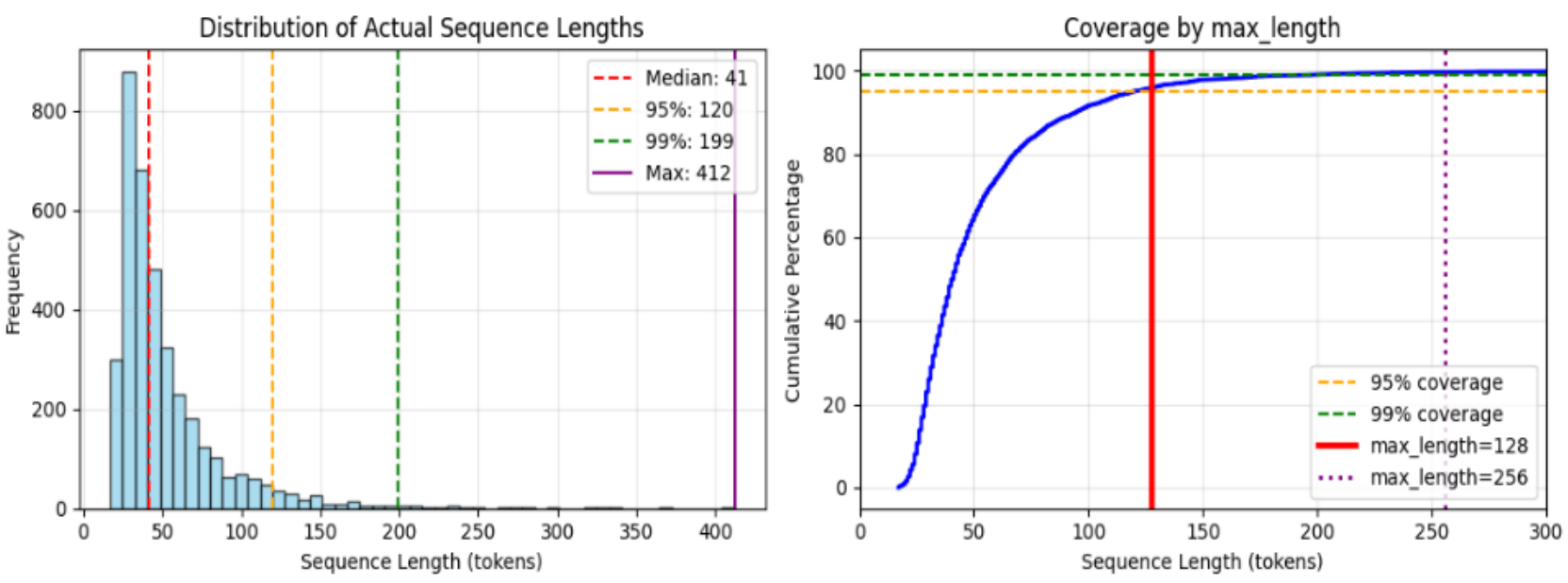

**Fig. 2.** Sequence length distribution and coverage analysis for 128-token threshold.

**Table 3.** Sequence construction

| <bos><problem>problem_text<user>user_text<user_emotion>emotion<br><therapist>therapist_text<therapist_emotion>emotion<eos> |
|---|

**Label Creation.** To train the model for response prediction, a selective labelling strategy masked context tokens (<problem>, <user>, <user_emotion>) and their content with a mask of -100. This methodology employed HuggingFace's Transformers library, where mask token -100 creates an “ignore index” without contributing to gradient loss. The response section employed differentiated labelling: whilst <therapist> markers were masked (-100), therapist text and <therapist_emotion> tokens received standard labels for prediction. This enabled response generation, including emotional state prediction, allowing the generation of content and emotional classification during inference.

# 4 Methodology

## 4.1 Experiment Setup

All experiments were conducted on Google Colab using an NVIDIA A100 GPU (40GB RAM). The baseline GPT-2 model (124M parameters) was imported via HuggingFace Transformers. RL used TRL 0.11.3 with PPO. Training was monitored using TensorBoard for SFT and Weights & Biases for RL experiments.

## 4.2 Baseline GPT-2

A retokenizer adapted GPT-2 to the structured input format (Table 3) for fair comparison.

### 4.3 Supervised Fine-tuning (With Therapist Emotions)

SFT used batches of 32 sequences (128 tokens) with AdamW optimizer (lr=2e-5), and cross-entropy loss. The model processed context-emotion unified inputs to establish a baseline for clinical dialogue generation.

### 4.4 Supervised Fine-tuning (No Therapist Emotions)

An ablation study excluded <therapist_emotion> tokens to quantify the value of joint text-emotion generation, producing only therapist text without emotional classification. This model helped quantify the joint text-emotion generation task's contribution to dialogue quality. The same hyperparameters were used as SFT (with emotions).

### 4.5 Supervised Fine-tuning + Reinforcement Learning

This study adapted the standard RL framework consisting of contexts C, response generation actions G, a policy π, and multi-dimensional rewards R. In this framework, given a context c ∈ C, the model generated a response g ∈ G according to the policy π: C → P(G), where P(G) represented the probability distribution over possible responses. RL was implemented on top of SFT according to Stiennon et al. 's [10] classical RL approach that utilised SFT as a stable starting policy and established basic task actions.

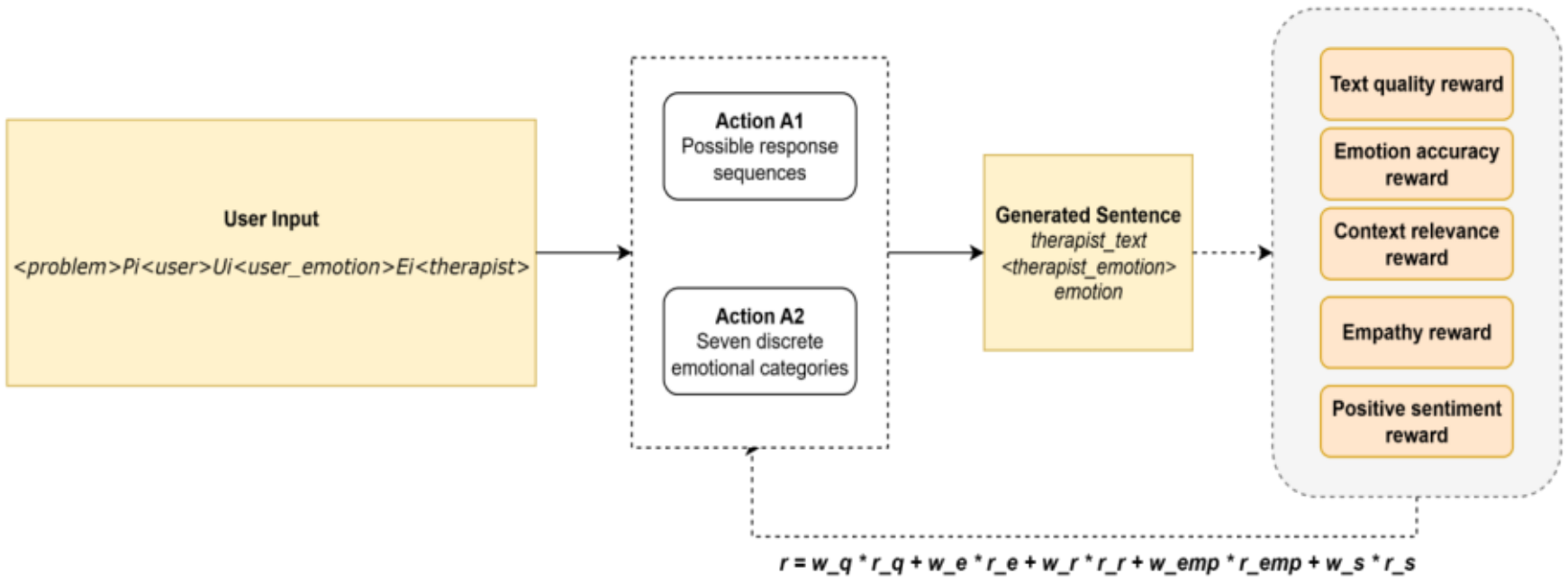

**Fig. 3.** The overall architectural representation of the reinforcement learning framework.

By employing a KL divergence penalty in RL, this approach prevented the RL policy from drifting too far from the SFT policy. Additionally, pre-trained token distributions may skew policy updates towards frequent words regardless of rewards. To mitigate this, this study integrated reward scaling to a range of -10 to 10 into the reward function [28]. A learning rate of $1 \times 10^{-6}$, training epochs of 8, batch size of 16, and data shuffling disabled were employed. Fig. 3 presents the architectural representation of the reinforcement learning framework.

**Proximal Policy Optimization (PPO).** PPO is a policy gradient method addressing training instability in reinforcement learning by constraining policy updates through a clipped objective function. The algorithm enables multiple epochs of minibatch up-dates whilst preventing excessive parameter changes that could destabilise training [29]. This stability was crucial for this study, where catastrophic forgetting posed risks to response patterns acquired during SFT.

The PPO objective optimised the clipped surrogate loss [see Eq. (1) and Eq. (2)]:

$$L^{CLIP(\theta)} = \hat{E}_t\ [\min\big(r_t(\theta)\hat{A}_t, clip(r_t(\theta), 1-\varepsilon, 1+\varepsilon)\hat{A}_t\big)] \quad (1)$$

where,

$$r_t(\theta) = \frac{\pi_\theta(a_t|s_t)}{\pi_{\theta_{old}}(a_t|s_t)} \quad (2)$$

represents the probability ratio between current and previous policies, $\hat{A}_t$denoted the advantage estimate that measures how much better an action is compared to the expected value at that state, and $\varepsilon$ was the clipping parameter that constrained policy updates within a trust region.

**Context.** The context enabled response generation specific to the patient's presenting issues, utterances, and emotional states at each dialogue turn. The context space consisted of structured user input processed through retokenisation. At each generation step, the context for each user utterance was represented by $C_i$, containing: problem type $P_i$, user utterance $U_i$, and user emotional state $E_i$, denoted by:

$$C_i =< problem > P_i < user > U_i < user_{emotion} > E_i$$

The $Ci$ is followed by the response prompt <therapist>, creating the complete input representation of context as $C_i < therapist >$

**Action.** The model executed actions through structured response generation, producing both therapist textual responses and corresponding emotional classifications. The action aimed to generate contextually appropriate responses recognising context $C_i$ whilst aligning with appropriate emotional classifications.

The model executed two generation components: $a1$ produced response text, and $a2$ predicted the appropriate therapist emotional stance. Action space $A1$ consisted of all possible response sequences within vocabulary constraints. Action space $A2$ encompassed seven discrete emotional categories existing within the dataset: {anger, sadness, depression, disgust, fear, joy, neutral}. The complete action was denoted as $a = (a1, a2) \in A1 * A2$. Following $C_i$, structured action generated responses in the sequence $< therapist_{emotion} > emotion_{word} < eos >$.

**Policy.** The policy operated by encoding structured input context $Ci$, then generating text by computing probability distributions over vocabulary tokens through decoder layers that maximized $P(response|context)$. The model continued generation beyond text to produce the emotional alignment sequence. Generation employed sampling parameters (top-p=1.0, top-k=0.0) recommended by TRL documentation to ensure response stability, with sequence completion triggered by the end-of-sequence token.

**Rewards.** Our novel composite reward function [Eq. (3)] addresses the limitation of prior RL models' overreliance on lexical metrics through complementary metrics: text quality, emotional alignment, contextual relevance, empathetic content, and sentiment appropriateness. The ensemble approach further addressed reward hacking, where models exploit individual reward functions by generating responses that achieve high scores through rule-following whilst producing clinically meaningless content [30].

**Text Fluency.** The text quality reward function $f_{quality}(\cdot)$ measured response coherence and appropriateness.

$$r_q = f_{quality}(therapist_{response})$$

Implementation incorporated repetition detection and meaningless pattern identification, penalising responses with excessive n-gram repetition and incoherent content.

**Emotional Alignment.** Emotional alignment rewarded suitable emotional classifications and penalised misalignment. The reward $r_e$ evaluated emotion prediction accuracy and emotion token presence:

$$r_e = f_{emotion}(predicted_{emotion}, target_{emotion}, has_emotion_token)$$

This rewarded exact emotional matches and appropriate categories (positive, negative, neutral), whilst penalising conflicts between incompatible categories.

**Contextual Relevance.** Contextual relevance ensured responses addressed specific patient concerns rather than generic interventions. Semantic similarity measured contextual appropriateness:

$$r_r = cos_sim(embed(therapist_response), embed(user_input))$$

Semantic embeddings were computed using Sentence-BERT (all-MiniLM-L6-v2) (Appendix 1) to measure similarity between responses and user inputs. This model was chosen for its ability to vectorise text for semantic comparison.

**Empathetic Content.** A pre-trained empathy classification model (paragon-analytics/bert_empathy) (Appendix 1) evaluated the empathetic quality of generated responses:

$$r_{emp} = f_{emp}(therapist_response)$$

The empathy model was chosen for its ability to predict empathic language content. This assessed empathetic quality, rewarding responses demonstrating understanding and emotional support.

**Sentiment Appropriateness.** Sentiment ensured responses conveyed appropriately positive sentiment, distinct from the discrete emotional categories used elsewhere. While emotion classification captured specific emotional states, sentiment analysis measured the overall positive/negative nature of the response

$$r_s = f_{sent}(therapist_response)$$

where, $f_{sent}(\cdot)$ employed DistilBERT sentiment classification (distilbert-base-uncased-finetuned-sst-2-english) (Appendix 1).

**Total Reward.** The final composite reward function combined all components with optimised weighting:

$$r_{total} = w_q \times r_q + w_e \times r_e + w_r \times r_r + w_{emp} \times r_{emp} + w_s \times r_s \quad (3)$$

where, text quality ($w_q$ = 1.1) and emotional alignment ($w_e$ = 1.2) received higher weighting, contextual understanding ($w_r$ = 1.1) was emphasised, whilst supportive content components ($w_{emp}$ = 0.7, $w_s$ = 0.7) received moderate weighting to balance clinical appropriateness with empathetic responsiveness.

Weight selection was based on preliminary experiments evaluating response quality across different configurations. The weights do not sum to 1.0 (total = 4.8) as this approach prioritised absolute reward magnitudes rather than relative proportions, where stronger gradients were needed to overcome the conservative learning rate ($1 \times 10^{-6}$) used for stability.

### 4.6 Hyperparameter Tuning

Grid search optimised top_p, top_k, and temperature parameters over combined BLEU and ROUGE-1-2-L metrics. This optimisation balanced response diversity and clinical appropriateness across the three model variants: SFT (No emotion), SFT (With emotion), and RL-SFT (With emotion). Full tuning parameters are provided in Appendix 2 (Table 9).

### 4.7 Evaluation

Evaluation used automated metrics as follows: LLM-as-a-judge on clinical criteria, and human evaluation-ensuring assessment aligns with clinical utility, not just NLP performance [27].

1. BLEU: precision against reference responses [31].
2. ROUGE-1-2-L: N-gram overlap between generated and reference responses [32].

3. METEOR: Automatic metric for machine translation evaluation based on generalised unigram matching [33].

LLM-as-a-judge assessed all three model variants using GPT-4o-mini via OpenAI API with prompt engineering (Table 10 in Appendix 3) to evaluate 100 samples from the testing dataset on a 1-5 scale across six emotional support criteria:

1. Therapeutic Rapport - Emotional connection and trust-building.
2. Active Understanding - Comprehension and reflection of user's state.
3. Relevance Focus - Addressing specific problems without going off-topic.
4. Practical Helpfulness - Providing actionable guidance and direction.
5. Professional Appropriateness - Maintaining boundaries.
6. Emotional Validation - Acknowledging and validating user emotions.

These criteria were adapted from Yuan et al.'s [34] human evaluation metrics to create a comprehensive framework tailored for therapeutic communication assessment. Human validation was conducted through manual inspection of a subset of GPT-4o-mini evaluations to ensure scoring consistency and appropriateness of the automated judgments.

# 5 Results

## 5.1 Automated Evaluation

Automated evaluation results (see Table 4) from baseline GPT-2 to the reinforcement learning fine-tuned model demonstrate substantial improvements across BLEU, ROUGE-1, ROUGE-2, ROUGE-L, and METEOR measures. Baseline GPT-2 achieved minimal scores across BLEU (0.0021) and ROUGE metrics (0.0478, 0.0064, 0.0408), indicating poor alignment with reference responses. Interestingly, baseline GPT-2 achieved the highest METEOR score (0.0670) across all models, likely due to METEOR's emphasis on synonym and stem matching, which may favour more diverse vocabulary generation over precise alignment with references [33].

**Table 4.** Automated evaluation quality metrics

| **Model** | **BLEU** | **ROUGE-1** | **ROUGE-2** | **ROUGE-L** | **METEOR** |
|---|---|---|---|---|---|
| GPT-2 | 0.0021 | 0.0478 | 0.0064 | 0.0408 | 0.0691 |
| SFT (No therapist emotions) | 0.0106 | 0.1160 | 0.0159 | 0.1058 | 0.0433 |
| SFT (Therapist emotions) | 0.0108 | 0.1164 | 0.0175 | 0.1076 | 0.0485 |
| Reinforcement Learning - SFT | 0.0111 | 0.1397 | 0.0213 | 0.1317 | 0.0581 |

Both SFT variants demonstrated improvements over baseline across BLEU and ROUGE metrics, though with reduced METEOR scores. The emotion-inclusive SFT model outperformed the no-emotion variant across most metrics, with notable improvements in ROUGE-2 (+10.1%) and ROUGE-L (+1.7%).

The RL-aligned model achieved the strongest overall performance, demonstrating substantial gains in ROUGE-1 (+20.0% over emotion SFT) and ROUGE-L (+22.4% over emotion SFT), suggesting enhanced lexical overlap with reference responses.

**Table 5.** Automated evaluation quality metrics (ESConv dataset)

| Model | BLEU | ROUGE-1 | ROUGE-2 | ROUGE-L | METEOR |
|---|---|---|---|---|---|
| SFT (Therapist emotions) | 0.0050 | 0.1046 | 0.0089 | 0.0925 | 0.0354 |
| Reinforcement Learning - SFT | 0.0042 | 0.1110 | 0.0092 | 0.1002 | 0.0290 |

In Table 5, automated evaluation on the ESConv dataset assessed model generalisation beyond the training domain. The SFT emotion-aware model achieved moderate cross-dataset performance. The RL-aligned model demonstrated mixed results, with improvements in ROUGE metrics (+6.1% ROUGE-1, +3.4% ROUGE-2, +8.3% ROUGE-L) compared to emotion-inclusive SFT, but decreased performance in BLEU (-16%) and METEOR (-18.1%).

**Table 6.** Emotion token generation accuracy performance

| Emotion Accuracy | Model |
|---|---|
| 0.6696 | GPT-2 |
| 0.8084 | SFT (No therapist emotions) |
| 0.9229 | SFT (Therapist emotions) |
| 0.9934 | Reinforcement Learning |

In Table 6, emotion classification accuracy revealed progressive improvements across model variants. Baseline GPT-2 achieved 66.96% accuracy, whilst the no-emotion SFT model reached 80.84%. The emotion-aware SFT model achieved 95.37%, demonstrating the effectiveness of joint text-emotion training. The RL model attained 99.34% accuracy, representing a 4.16% improvement over emotion-aware SFT. This high performance demonstrates that the composite reward function effectively enhanced emotional alignment capabilities.

## 5.2 LLM-as-a-Judge Evaluation

In Table 7, LLM-as-a-judge evaluation (GPT-4o-mini) revealed a clear performance hierarchy across all models tested on 100 dialogue samples. The RL model achieved the highest overall average score of 1.718, followed by SFT at 1.655, and baseline GPT-2 at 1.062.

The RL model demonstrated notable superiority in relevance focus (1.830 vs SFT's 1.740), representing 5.2% improvement. This aligns with the contextual relevance component in the composite reward function, suggesting successful optimisation of context-aware response generation. In active understanding, the RL model achieved 1.670 compared to SFT's 1.570 (6.4% improvement), indicating enhanced comprehension of patient concerns. In emotional validation, RL achieved 1.640 versus SFT's 1.530 (7.2% improvement), validating the emotion-focused reward components.

However, both advanced models showed similar performance in practical helpfulness (RL: 1.490, SFT: 1.480), with only marginal improvement over baseline GPT-2 (1.020).

**Table 7.** LLM-as-a-judge evaluation results

| Metric | Baseline GPT-2 | SFT Model | RL Model |
|---|---|---|---|
| Therapeutic Rapport | 1.090 | 1.700 | 1.730 |
| Active Understanding | 1.030 | 1.570 | 1.670 |
| Relevance Focus | 1.070 | 1.740 | 1.830 |
| Practical Helpfulness | 1.020 | 1.480 | 1.490 |
| Professional Appropriateness | 1.100 | 1.910 | 1.950 |
| Emotional Validation | 1.060 | 1.530 | 1.640 |

### 5.3 Human Evaluation

Manual inspection (samples in Table 8) revealed that LLM-as-a-judge had under-scored around 20 samples. Human evaluation showed that the RL model frequently generated question-oriented responses rather than providing direct therapeutic advice. For instance (Table 8), when users express intense emotions such as depression regarding bereavement, the generated responses employ open-ended questioning techniques that encourage further exploration without emotional reactivity. However, this likely explains the relatively low ROUGE scores, as reference responses were derived from conversations characterised by natural turn-taking, where not every utterance tends to be a question. Consequently, there was lower lexical overlap with reference responses that contained more declarative statements.

Nevertheless, generated responses demonstrated strong contextual relatedness, with every response directly addressing specific keywords in patient utterances rather than providing generic statements. The neutral emotions generated benefit therapeutic rapport. According to Gelso and Kanninen [22], effective therapists maintain neutrality to remain close to the patient's “zone of proximal development”, avoiding manipulation of patient emotional expression through excessive warmth. The RL model's tendency towards neutral emotional stances and seemingly low Emotional Validation scores reflects this professional therapeutic approach.

**Table 8.** Samples of generated responses from the SFT-RL model

| **Problem Type** | **User Text** | **User emotion** | **Generated Therapist Response** | **Generated Therapist Emotion** |
|---|---|---|---|---|
| Therapeutic Relationship | Did I offend you? | neutral | No, you didn't offend me. You didn't say that you were upset with me. | neutral |
| Bereavement | I can't stop thinking about my grandmother who passed away | depression | And you wanna tell me what you wanna do for her? | neutral |
| Post-Traumatic Stress Disorder | I keep having flashbacks from the accident | fear | Do you want to talk about it? | neutral |

# 6 Discussion

This study presents a proof-of-concept for developing cost-effective, locally deployable therapeutic dialogue models using structured multimodal input formatting in GPT-2 combined with multi-component reward optimisation. The approach addresses critical requirements for mental health applications: data sensitivity through local deployment, computational efficiency for resource-constrained environments, and enhanced safety through smaller model parameters that enable better oversight and control [20]. Our results confirm that the proposed approach directly addresses the limitations of existing methods:

1. **Clinical Empathy Alignment**: Unlike SFT models [24] that produce surface empathy, our RL model's 99.34% emotion accuracy and strong LLM-as-a-judge scores in Emotional Validation demonstrate that integrating clinical-specific rewards (emotional alignment, contextual relevance) generates empathy that is both genuine and boundary-appropriate. This addresses Luo et al.'s [13] finding that patients prefer human responses-our model's empathy is tied to context, not scripted.
2. **Context-Emotion Integration**: The RL model's superior performance on Relevance Focus (+5.2% over SFT) and cross-dataset ROUGE improvements confirm that structuring input as unified context-emotion units (vs. separate features) solves prior models' context-insensitivity. Unlike Saha et al. [16], our model seeks not to generate generic empathy phrases – every response addresses specific user keywords.
3. **Reward Function Superiority**: Our multi-component reward function outperforms generic metrics (e.g., BLEU, ROUGE) by prioritising clinical utility. The RL model's higher Professional Appropriateness scores (1.95) demonstrate that rewarding therapeutic boundaries avoids the over-empathising trap of prior RL models [35].

Notably, these responses are generated upon GPT-2 (124M parameters), a decoder-only model with inherent architectural limitations for complex conversational tasks. Researchers from Anthropic found that 'alignment tax' is inherent in RL training on small LLMs (<10B parameters) where NLP metrics performance may degrade after RL training [36].

Comparing against existing empirical evidence, Sharma et al. [15] performed RL on DialoGPT for empathetic rewriting, transforming low-empathy conversational posts to higher empathy. Whilst serving different purposes, it demonstrated RL's feasibility in improving response generation, achieving a BLEU score of 0.1391. Although this score is significantly higher than our model's, this does not diminish our results as their task differs fundamentally from dialogue generation. Empathetic rewriting resembles machine translation – improving existing text rather than generating new responses - creating an inherent advantage when using BLEU, a metric designed to evaluate machine-translated text against reference translations.

Evaluating Saha et al.'s [16] work, which performed RL on GPT-2 using composite rewards of BLEU, ROUGE-L, and sentiment scores for empathetic text generation, this study recognises that such metrics may prioritise lexical similarity over contextual appropriateness and often yield sparse, biased signals [28]. Whilst direct performance comparison is not meaningful due to different datasets and tasks, our approach demonstrates that composite rewards incorporating contextual and emotional appropriateness can achieve competitive lexical scores (ROUGE-L: 0.1317) without directly optimising for them.

An important aspect of this model is improving current therapist generation approaches by enhancing affective empathy, where SFT-emotion outperformed SFT-no-emotion across all metrics (BLEU: 0.0108 vs. 0.0106; ROUGE-1: 0.1164 vs. 0.1160; ROUGE-2: 0.0175 vs. 0.0159; ROUGE-L: 0.1076 vs. 0.1058; METEOR: 0.0485 vs. 0.0433). This suggests that generating emotion tokens before end-of-sentence tokens provides dual benefits: improving lexical generation whilst demonstrating affective empathy, which Rudra et al. [35] emphasised as important for therapists to display genuine emotional engagement rather than surface-level recognition.

This work's contribution lies in demonstrating a proof-of-concept for developing smaller, safer therapeutic dialogue models suitable for sensitive healthcare data. By restructuring GPT-2 for contextual-emotional input processing, we present a cost-effective multimodal alternative that addresses critical deployment constraints in mental health applications: data privacy through local deployment, computational efficiency for resource-limited environments, and enhanced oversight capabilities inherent in smaller model architectures. The SFT model's consistent improvements across automated evaluation metrics, validated by LLM-as-a-judge evaluation showing 'Relevance Focus' as the second-highest metric, demonstrates that structured input effectively captures context-emotion relationships without requiring large-scale, resource-intensive models. This approach offers healthcare institutions a viable pathway for implementing AI-assisted therapeutic tools whilst maintaining stringent data security and clinical supervision requirements.

### 6.1 Limitations

This study acknowledges several important limitations. Firstly, significant class imbalance in emotion labels within the training dataset resulted in overfitting towards neutral therapist emotion generation. This creates bias where the model defaults to emotionally restrained responses, potentially limiting communication flexibility. Whilst this reflects realistic professional therapist approaches to maintaining clinical boundaries, it may reduce empathy and emotional responsiveness in contexts requiring warmer interactions. This limitation appears in LLM-as-a-judge evaluation results, where the model achieved the lowest score for Practical Helpfulness whilst rating highest in Professional Appropriateness, suggesting focus on professional behaviour over empathetic engagement.

Another significant limitation is the absence of empirical hyperparameter optimisation for reward weights prior to training. Given limited computational resources and time, systematic optimisation techniques were omitted. Human evaluation reveals the relevance component may be disproportionately weighted, leading to suboptimal conversational patterns where the model occasionally uses simplistic paraphrasing rather than generating meaningful responses. For instance, when presented with "My ex keeps texting me, and I don't know how to respond", the model sometimes produces repetitive responses like "So you're not sure how to respond?" This suggests excessive relevance weighting may encourage surface-level lexical matching rather than deeper engagement.

Significant limitations exist in using LLMs for mental health support. First, while these models simulate empathy effectively, they frequently exaggerate it through lengthy responses, creating misleading emotional connections and potentially fostering inappropriate user reliance [35] [37]. Second, generic prompt engineering fails to capture contextual awareness, resulting in inconsistent responses that undermine reliability [11] [38]. Third, LLMs lack clinical judgement, making them prone to generating harmful recommendations in legally sensitive psychological contexts [39]. Fourth, pre-trained models exhibit racial and demographic biases, producing stereotyped responses and hallucinated inaccuracies that misalign with users' values [21] [12]. Finally, commercially available LLMs raise privacy concerns regarding API vulnerabilities and data leakage when handling sensitive information [40].

### 6.2 Future Research

Whilst this study employed a relatively small LLM with only 124M parameters, qualitative examples demonstrated GPT-2's capability to produce responses that effectively capture context and remain highly relevant to user input. Fine-tuning on therapy session data shows its effectiveness in enabling LLMs to replicate professionalism and appropriate speech tones, potentially addressing limitations of generalised LLMs that produce superficial empathetic responses through lengthy, generic outputs. This study highlights the critical need for higher-quality datasets to better align LLMs with human needs and serve vulnerable groups requiring specialised support.

The structured input approach demonstrated success and reveals potential for improving emotional support CAs to capture multimodal information. Whilst multimodality here involves preprocessing video and audio signals into text, future research could employ RL on graph-convolutional-network-based multimodal classifiers [41], restructuring input formats for multimodal response generation.

# 7 Conclusion

This study directly addresses four critical methodological gaps in existing therapeutic dialogue generation: 1) SFT's failure to balance clinical rigor and empathy; 2) RL's overreliance on generic reward metrics; 3) context-emotion decoupling; and 4) deployment barriers. Our innovations – a clinical-specific multi-component reward function, an RL framework for context-emotion integration, and a lightweight GPT-2 implementation – produce a model that achieves 99.34% emotion accuracy, outperforms baseline and SFT variants on clinical criteria, and enables local deployment. The results demonstrate that small LLMs, when optimised for clinical priorities (not just NLP metrics), can serve as valuable assistive tools for therapists – enhancing accessibility to mental health support while maintaining clinical safety and privacy. This work paves the way for practical, ethical AI-assisted therapeutic tools that address the global mental health burden.

# 8 Ethics

All datasets are open source from published articles. Recognising potential harms of LLMs in mental health support [42], including retraumatisation, unreliable advice, overreliance, and safety bypass risks, this study positions the model as a clinical assistance tool rather than a direct patient interface. Implementation follows a tiered ethical framework [43]: Level 1 Safeguards (expert-use disclaimers, local deployment), Level 2 Protections (human oversight, transcript review), and Level 3 Responsibility (transparent documentation of limitations and capabilities).

**Declaration on Generative AI.** GenAI has been used for supporting drafting and language refinement.

# Appendices

### Appendix 1: URLs for Tools

Tensorboard: https://www.tensorflow.org/tensorboard
HuggingFace Transformers library https://huggingface.co/docs/transformers/en/index
HuggingFace Transformers GPT2 https://huggingface.co/openai-community/gpt2
HuggingFace TRL Reinforcement Learning PPO:
https://huggingface.co/docs/trl/main/en/ppo_trainer
https://pypi.org/project/trl/0.11.3/
Weights & Biases: https://wandb.ai/site/
roberta-base-go_emotions:
https://huggingface.co/SamLowe/roberta-base-go_emotions
all-MiniLM-L6-v2: https://huggingface.co/sentence-transformers/all-MiniLM-L6-v2
distilbert-base-uncased-finetuned-sst-2-english:
https://huggingface.co/distilbert/distilbert-base-uncased-finetuned-sst-2-english

paragon-analytics/bert_empathy:
https://huggingface.co/paragon-analytics/bert_empathy

**Appendix 2**

**Table 9.** Hyperparameters

| Model | Range |
| --- | --- |
| GPT-2 | TOP_P_VALUES = [0.6, 0.65, 0.7, 0.75, 0.8, 0.85, 0.9, 0.95, 1.00]<br>TOP_K_VALUES = [0, 5, 10, 15, 20, 25, 30, 35]<br>TEMPERATURE_VALUES = [1.0, 1.1, 1.2, 1.3] |
| SFT (With emotions) | TOP_P_VALUES = [0.6, 0.65, 0.7, 0.75, 0.8, 0.85, 0.9, 0.95, 1.00]<br>TOP_K_VALUES = [0, 5, 10, 15, 20, 25, 30, 35]<br>TEMPERATURE_VALUES = [1.0, 1.1, 1.2, 1.3] |
| Reinforcement Learning - SFT (With emotions) | TOP_P_VALUES = [0.6, 0.65, 0.7, 0.75, 0.8, 0.85, 0.9, 0.95, 1.00]<br>TOP_K_VALUES = [0, 5, 10, 15, 20, 25, 30, 35]<br>TEMPERATURE_VALUES = [1.0, 1.1, 1.2, 1.3] |

**Appendix 3**

**Table 10.** Prompt for LLM-as-a-judge Evaluation

| Criterion | Description | Criteria examples |
| --- | --- | --- |
| Therapeutic Rapport | Evaluates emotional connection, compassion, active listening skills, and ability to provide affirmation and comfort to build trust with the user | Look for: Warmth, validation, emotional attunement, supportive tone, building trust<br>Poor examples: Dismissive, cold, judgmental responses |
| Active Understanding | Measures comprehension of the user's emotional state and situation through effective paraphrasing, reflection, and demonstration of understanding | Look for: Accurate reflection of user's words, paraphrasing key points, demonstrating comprehension of the situation<br>Poor examples: Misunderstanding, ignoring context |
| Relevance Focus | Assesses how well the response addresses the specific problem or situation mentioned by the user without going off-topic | Look for: Directly addressing the stated problem, staying on topic, responding to user's specific concerns<br>Poor examples: Tangential responses, changing subjects |
| Practical Helpfulness | Evaluates whether the response provides practical value, support, and clear therapeutic direction to help the user progress | Look for: Actionable guidance, therapeutic techniques, constructive suggestions, movement toward solutions<br>Poor examples: Vague responses, no direction |
| Professional Appropriateness | Measures appropriate therapeutic interpretation of statements and judicious use of self-disclosure while maintaining professional boundaries | Look for: Proper therapeutic boundaries, appropriate interpretations, professional language and approach<br>Poor examples: Oversharing, inappropriate interpretations |
| Emotional | Assesses the therapist's abil- | Look for: Acknowledging user's feelings, |

| Validation | ity to acknowledge, validate, and appropriately respond to the user's current emotional state | normalizing emotions, showing understanding of emotional experience<br>Poor examples: Dismissing feelings, minimizing emotions |
|---|---|---|

**prompt** = f"""You are an expert evaluator of therapy chatbot responses. Your task is to evaluate a therapist chatbot's response based on the specific criterion provided.

**Context:**

1. Problem Type: {problem_type} # Type of user's psychological problem (e.g., anxiety, depression)
2. User Input: "{user_text}" # Original input content from the user
3. User Emotion: {user_emotion} # Emotional state of the user (e.g., frustrated, anxious)
4. Therapist Response: "{therapist_response}" # Output from the therapy chatbot

**Evaluation Criterion: {criteria}**
Description: {self.evaluation_criteria.get(criteria, "General therapeutic effectiveness")} # Criterion definition from the evaluation table
Evaluation Focus: {criteria_examples.get(criteria, "Evaluate overall therapeutic quality.")} # Key observation points for the criterion

**Instructions:**

1. Evaluate the therapist response on a scale of 1-5 for the given criterion:

 - 1: Poor - Completely inadequate or inappropriate for this criterion
 - 2: Below Average - Some issues or limitations in this specific area
 - 3: Average - Adequate performance in this criterion but room for improvement
 - 4: Good - Well-executed in this criterion with minor areas for enhancement
 - 5: Excellent - Outstanding and highly effective in this specific criterion

2. Provide a focused explanation (2-3 sentences) specifically about this criterion.
3. Consider how well the response performs in this particular dimension of therapeutic effectiveness.

**Response Format:**
Score: [1-5]
Explanation: [Your justification focused specifically on this criterion]

**Your Evaluation:**"""

## References

1. 'Mental Illness - National Institute of Mental Health (NIMH)'. Accessed: Sep. 13, 2025. [Online]. Available: https://www.nimh.nih.gov/health/statistics/mental-illness

2. P. Corrigan, 'How stigma interferes with mental health care.', Am. Psychol., vol. 59, no. 7, pp. 614–625, Oct. 2004, doi: 10.1037/0003-066X.59.7.614.
3. B. M. Coêlho, G. L. Santana, M. C. Viana, Y.-P. Wang, and L. H. Andrade, "I don't need any treatment" – barriers to mental health treatment in the general population of a megacity', Braz. J. Psychiatry, vol. 43, no. 6, pp. 590–598, Dec. 2021, doi: 10.1590/1516-4446-2020-1448.
4. J. Goodwin, E. Savage, and A. O'Donovan, "I Personally Wouldn't Know Where to Go": Adolescents' Perceptions of Mental Health Services', J. Adolesc. Res., vol. 39, no. 5, pp. 1384–1412, Sep. 2024, doi: 10.1177/07435584221076056.
5. S. Li, M. Chen, P. L. Liu, and J. Xu, 'Following Medical Advice of an AI or a Human Doctor? Experimental Evidence Based on Clinician-Patient Communication Pathway Model', Health Commun., vol. 40, no. 9, pp. 1810–1822, Jul. 2025, doi: 10.1080/10410236.2024.2423114.
6. E. Jo et al., 'Assessing GPT-4's Performance in Delivering Medical Advice: Comparative Analysis With Human Experts', JMIR Med. Educ., vol. 10, no. 1, p. e51282, Jul. 2024, doi: 10.2196/51282.
7. A. Følstad et al., 'Future directions for chatbot research: an interdisciplinary research agenda', Computing, vol. 103, no. 12, pp. 2915–2942, Dec. 2021, doi: 10.1007/s00607-021-01016-7.
8. D. O. Shumway and H. J. Hartman, 'Medical malpractice liability in large language model artificial intelligence: legal review and policy recommendations', J. Osteopath. Med., vol. 124, no. 7, pp. 287–290, Jul. 2024, doi: 10.1515/jom-2023-0229.
9. C. Raffel et al., 'Exploring the Limits of Transfer Learning with a Unified Text-to-Text Transformer'.
10. N. Stiennon et al., 'Learning to summarize with human feedback', in Advances in Neural Information Processing Systems, Curran Associates, Inc., 2020, pp. 3008–3021. Accessed: Apr. 08, 2025. [Online]. Available: https://proceedings.neurips.cc/paper_files/paper/2020/hash/1f89885d556929e98d3ef9b86448f951-Abstract.html
11. Z. Guo, A. Lai, J. H. Thygesen, J. Farrington, T. Keen, and K. Li, 'Large Language Model for Mental Health: A Systematic Review', Feb. 18, 2024. doi: 10.2196/preprints.57400.
12. L. Ouyang et al., 'Training language models to follow instructions with human feedback'.
13. M. Luo, C. J. Warren, L. Cheng, H. M. Abdul-Muhsin, and I. Banerjee, 'Assessing Empathy in Large Language Models with Real-World Physician-Patient Interactions', in 2024 IEEE International Conference on Big Data (BigData), Dec. 2024, pp. 6510–6519. doi: 10.1109/BigData62323.2024.10825307.
14. H. Lang, F. Huang, and Y. Li, 'Fine-Tuning Language Models with Reward Learning on Policy', Mar. 28, 2024, arXiv: arXiv:2403.19279. doi: 10.48550/arXiv.2403.19279.
15. A. Sharma, I. W. Lin, A. S. Miner, D. C. Atkins, and T. Althoff, 'Towards Facilitating Empathic Conversations in Online Mental Health Support: A Reinforcement Learning Approach', in Proceedings of the Web Conference 2021, Ljubljana Slovenia: ACM, Apr. 2021, pp. 194–205. doi: 10.1145/3442381.3450097.
16. T. Saha, V. Gakhreja, A. S. Das, S. Chakraborty, and S. Saha, 'Towards Motivational and Empathetic Response Generation in Online Mental Health Support', in Proceedings of the 45th International ACM SIGIR Conference on Research and Development in Information Retrieval, Madrid Spain: ACM, Jul. 2022, pp. 2650–2656. doi: 10.1145/3477495.3531912.
17. R. L. Bashshur, G. W. Shannon, N. Bashshur, and P. M. Yellowlees, 'The Empirical Evidence for Telemedicine Interventions in Mental Disorders', Telemed. E-Health, vol. 22, no. 2, pp. 87–113, Feb. 2016, doi: 10.1089/tmj.2015.0206.

18. E. J. Kraus, B. Nicosia, and D. I. Shalowitz, 'A qualitative study of patients' attitudes towards telemedicine for gynecologic cancer care', Gynecol. Oncol., vol. 165, no. 1, pp. 155–159, Apr. 2022, doi: 10.1016/j.ygyno.2022.01.035.
19. M. Torniainen-Holm et al., 'The effectiveness of email-based exercises in promoting psychological wellbeing and healthy lifestyle: a two-year follow-up study', BMC Psychol., vol. 4, no. 1, p. 21, Dec. 2016, doi: 10.1186/s40359-016-0125-4.
20. Z. Ma, Y. Mei, and Z. Su, 'Understanding the Benefits and Challenges of Using Large Language Model-based Conversational Agents for Mental Well-being Support', AMIA. Annu. Symp. Proc., vol. 2023, pp. 1105–1114, Jan. 2024.
21. S. Gabriel, I. Puri, X. Xu, M. Malgaroli, and M. Ghassemi, 'Can AI Relate: Testing Large Language Model Response for Mental Health Support', Oct. 07, 2024, arXiv: arXiv:2405.12021. doi: 10.48550/arXiv.2405.12021.
22. C. Gelso and K. Kanninen, 'Neutrality Revisited: On the Value of Being Neutral Within an Empathic Atmosphere', J. Psychother. Integr., vol. 27, pp. 330–341, Feb. 2017, doi: 10.1037/int0000072.
23. H. W. Chung et al., 'Scaling Instruction-Finetuned Language Models'.
24. H. Jin, S. Chen, D. Dilixiati, Y. Jiang, M. Wu, and K. Q. Zhu, 'PsyEval: A Suite of Mental Health Related Tasks for Evaluating Large Language Models', Jun. 03, 2024, arXiv: arXiv:2311.09189. doi: 10.48550/arXiv.2311.09189.
25. A. Radford, J. Wu, R. Child, D. Luan, D. Amodei, and I. Sutskever, 'Language Models are Unsupervised Multitask Learners'.
26. Y. Chu, L. Liao, Z. Zhou, C.-W. Ngo, and R. Hong, 'Towards Multimodal Emotional Support Conversation Systems', Oct. 19, 2024, arXiv: arXiv:2408.03650. doi: 10.48550/arXiv.2408.03650.
27. S. Liu et al., 'Towards Emotional Support Dialog Systems', Jun. 02, 2021, arXiv: arXiv:2106.01144. doi: 10.48550/arXiv.2106.01144.
28. A. Anuchitanukul and J. Ive, 'SURF: Semantic-level Unsupervised Reward Function for Machine Translation', in Proceedings of the 2022 Conference of the North American Chapter of the Association for Computational Linguistics: Human Language Technologies, M. Carpuat, M.-C. de Marneffe, and I. V. Meza Ruiz, Eds, Seattle, United States: Association for Computational Linguistics, Jul. 2022, pp. 4508–4522. doi: 10.18653/v1/2022.naacl-main.334.
29. J. Schulman, F. Wolski, P. Dhariwal, A. Radford, and O. Klimov, 'Proximal Policy Optimization Algorithms', Aug. 28, 2017, arXiv: arXiv:1707.06347. doi: 10.48550/arXiv.1707.06347.
30. J. Eisenstein et al., 'Helping or Herding? Reward Model Ensembles Mitigate but do not Eliminate Reward Hacking', Aug. 16, 2024, arXiv: arXiv:2312.09244. doi: 10.48550/arXiv.2312.09244.
31. K. Papineni, S. Roukos, T. Ward, and W.-J. Zhu, 'BLEU: a method for automatic evaluation of machine translation', in Proceedings of the 40th Annual Meeting on Association for Computational Linguistics - ACL '02, Philadelphia, Pennsylvania: Association for Computational Linguistics, 2001, p. 311. doi: 10.3115/1073083.1073135.
32. C.-Y. Lin, 'ROUGE: A Package for Automatic Evaluation of Summaries', in Text Summarization Branches Out, Barcelona, Spain: Association for Computational Linguistics, Jul. 2004, pp. 74–81. Accessed: Sep. 13, 2025. [Online]. Available: https://aclanthology.org/W04-1013/
33. S. Banerjee and A. Lavie, 'METEOR: An Automatic Metric for MT Evaluation with Improved Correlation with Human Judgments', in Proceedings of the ACL Workshop on Intrinsic and Extrinsic Evaluation Measures for Machine Translation and/or Summarization,

J. Goldstein, A. Lavie, C.-Y. Lin, and C. Voss, Eds, Ann Arbor, Michigan: Association for Computational Linguistics, Jun. 2005, pp. 65–72. Accessed: Sep. 13, 2025. [Online]. Available: https://aclanthology.org/W05-0909/

34. A. Yuan, E. Garcia Colato, B. Pescosolido, H. Song, and S. Samtani, 'Improving Workplace Well-being in Modern Organizations: A Review of Large Language Model-based Mental Health Chatbots', ACM Trans. Manag. Inf. Syst., vol. 16, no. 1, pp. 1–26, Mar. 2025, doi: 10.1145/3701041.
35. P. Rudra, W.-T. Balke, T. Kacprowski, F. Ursin, and S. Salloch, 'Large language models for surgical informed consent: an ethical perspective on simulated empathy', J. Med. Ethics, p. jme-2024-110652, Mar. 2025, doi: 10.1136/jme-2024-110652.
36. Y. Bai et al., 'Training a Helpful and Harmless Assistant with Reinforcement Learning from Human Feedback', Apr. 12, 2022, arXiv: arXiv:2204.05862. doi: 10.48550/arXiv.2204.05862.
37. V. Sorin et al., 'Large Language Models and Empathy: Systematic Review', J. Med. Internet Res., vol. 26, no. 1, p. e52597, Dec. 2024, doi: 10.2196/52597.
38. F. Farhat, 'ChatGPT as a Complementary Mental Health Resource: A Boon or a Bane', Ann. Biomed. Eng., vol. 52, no. 5, pp. 1111–1114, May 2024, doi: 10.1007/s10439-023-03326-7.
39. D. Grabb, 'The impact of prompt engineering in large language model performance: a psychiatric example', J. Med. Artif. Intell., vol. 6, no. 0, Oct. 2023, doi: 10.21037/jmai-23-71.
40. H. R. Lawrence, R. A. Schneider, S. B. Rubin, M. J. Matarić, D. J. McDuff, and M. J. Bell, 'The Opportunities and Risks of Large Language Models in Mental Health', JMIR Ment. Health, vol. 11, no. 1, p. e59479, Jul. 2024, doi: 10.2196/59479.
41. Q. Yang, M. Ye, and B. Du, 'EmoLLM: Multimodal Emotional Understanding Meets Large Language Models', Jun. 29, 2024, arXiv: arXiv:2406.16442. doi: 10.48550/arXiv.2406.16442.
42. I. Song, S. R. Pendse, N. Kumar, and M. D. Choudhury, 'The Typing Cure: Experiences with Large Language Model Chatbots for Mental Health Support', Mar. 06, 2024, arXiv: arXiv:2401.14362. doi: 10.48550/arXiv.2401.14362.
43. N. Neveditsin, P. Lingras, and V. Mago, 'Clinical Insights: A Comprehensive Review of Language Models in Medicine', Jan. 07, 2025, arXiv: arXiv:2408.11735. doi: 10.48550/arXiv.2408.11735.